\documentclass[11pt]{article}

\usepackage[utf8]{inputenc} 
\usepackage[T1]{fontenc}    
\usepackage{lmodern}
\usepackage{url}            
\usepackage{booktabs}       
\usepackage{amsfonts}       
\usepackage{nicefrac}       
\usepackage{microtype}      
\usepackage{float}
\usepackage{enumerate}

\usepackage{mathrsfs}
\usepackage{amsmath}
\usepackage{amssymb}
\usepackage{natbib}
\usepackage{graphicx}
\usepackage{url}
\PassOptionsToPackage{x11names}{xcolor}
\usepackage{xcolor}
\PassOptionsToPackage{algo2e,ruled}{algorithm2e}
\usepackage{algorithm2e}
\usepackage{jmlrutils}
\usepackage{hyperref}
\usepackage{nameref}

\usepackage{authblk}

\usepackage{pdfpages} 

\begin{document}
\title{\bf Double descent in the condition number}

\author{Tomaso Poggio, Gil Kur, Andrzej Banburski}
\affil{Center for Brains, Minds, and Machines, MIT}

\maketitle

\begin{abstract}
In solving a system of $n$ linear equations in $d$ variables
  $Ax=b$, the condition number of the $n,d$ matrix $A$ measures how
  much errors in the data $b$ affect the solution $x$. Estimates of
  this type are important in many inverse problems. An example is
  machine learning where the key task is to estimate an underlying
  function from a set of measurements at random points in a high
  dimensional space and where low sensitivity to error in the data is
  a requirement for good predictive performance. Here we discuss the
  simple observation, which is known but surprisingly little quoted (see
  Theorem 4.2 in \cite{Brgisser:2013:CGN:2526261}): when the columns of $A$ are random vectors,
  the condition number of $A$ is highest if $d=n$,
  that is when the inverse of $A$ exists. An overdetermined system
  ($n>d$) as well as an underdetermined system ($n<d$), for which
  the pseudoinverse must be used instead of the inverse, typically
  have significantly better, that is lower, condition numbers. Thus
  the condition number of $A$ plotted as function of $d$ shows a
  double descent behavior with a peak at $d=n$.
\end{abstract}

The concept of condition number was introduced by Turing in 1948
\cite{Turing48} and has since played a key role in the theory of
algorithms.  The condition number of a function measures how much the
output value of the function can change for a small change in the
input argument. The condition number most commonly associated with
$Ax=b$ is defined as the ratio of the relative error in $x$  to
the relative error in the data $b$. In terms of the $l_2$ norm on $x$
and $b$, this leads to the following definition for the the
condition number of $A$, denoted by $\kappa(A)=||A||||A^\dagger||$ with
$||A||$ being the operator norm of the $m,n$ matrix $A$ and $A^\dagger$
the pseudoinverse. The operator  norm is defined
as $||A||=\sup_{x}||Ax||$ with $||x||=1$. 
it is easy to see that
$\kappa(A) =\frac{\sigma_{max}(A)}{\sigma_{min}(A)}$ is the ratio
of the maximal and minimal singular values of $A$.

\begin{figure}
	\centering
	\includegraphics[trim = 115 240 125 255, width=1\linewidth, clip]{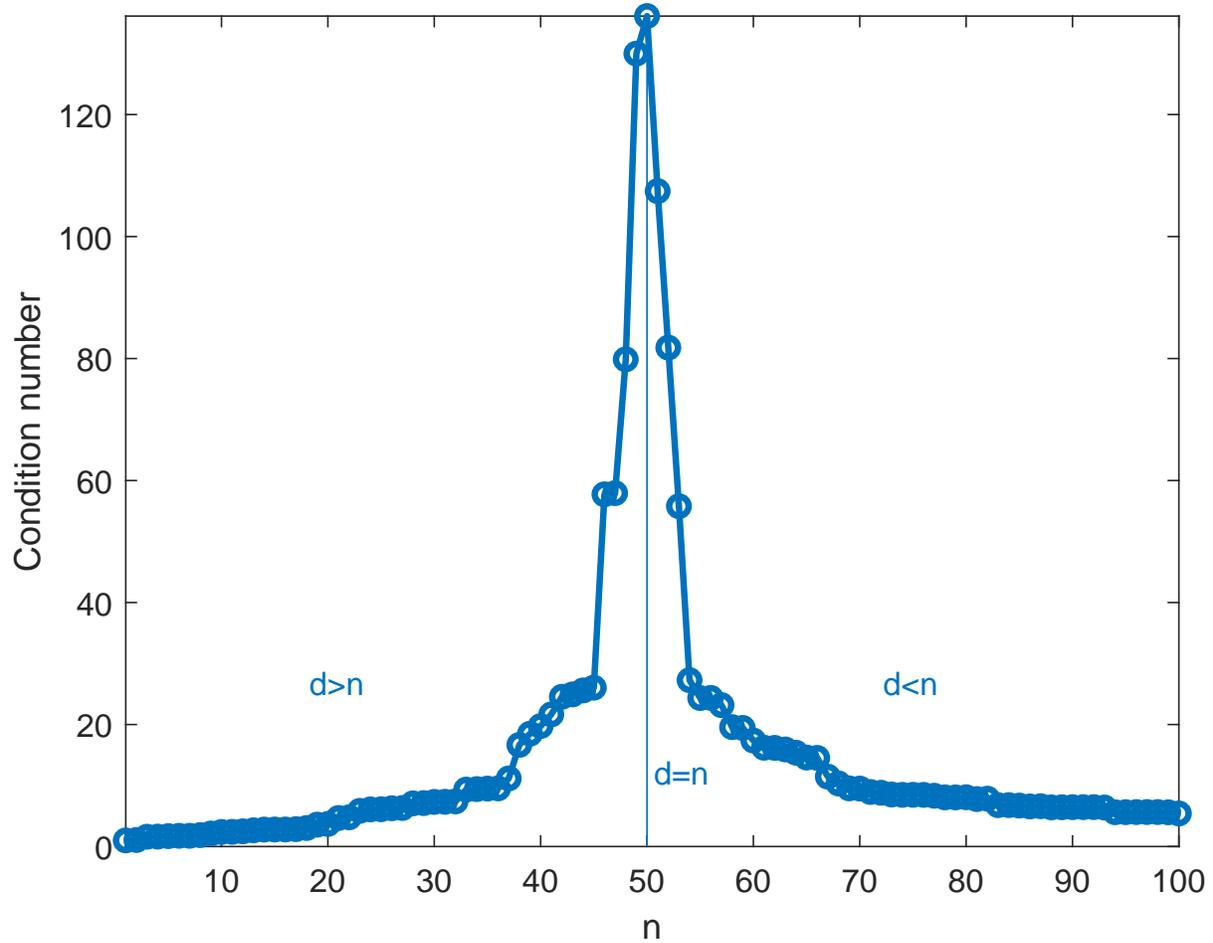}
	\caption{The typical ''double descent"  of
	 a $n \times d$ matrix with $\mathcal{N}(0,1)$ independent entries. The condition number is the worst when $n=d$. }
	\label{CondNumber}
\end{figure}

The plot in the Figure \ref{CondNumber} can be easily checked by calling the function
``cond'' in MatLab. The double descent pattern is apparently quite
robust to choices of $d$ and $n$, such that their ratio
$\gamma=\frac{n}{d}$ is the same. The fact that the worse conditioning
occurs when the inverse exists uniquely ($\gamma=1$) seems at first
surprising. This simple observation must have been realized by many.  The proof is also
simple because of a well-known characterization of the
eigenvalues of random matrices \cite{MarchenkoPastur}.  In fact,
consider a $n \times d$ random matrix $A$. 

We characterize its condition
number by using the Marchenko–Pastur semi-circle law, which describes the
asymptotic behavior of singular values of large rectangular random
matrices. We assume that the entries of $A$ are i.d.d. random variables with mean zero and variance
one. We consider the limit for $n \to \infty$ with
$\frac{n}{d} \to \gamma$. 

Marchenko–Pastur claims that for $\gamma <1$
the smallest and the largest singular values of $\frac{1}{d}AA^T$ are,
respectively $(1-\sqrt{\gamma})^2$ and $(1+\sqrt{\gamma})^2$. For
$\gamma >1$ the largest and the smallest eigenvalues of
$\frac{1}{n}A^TA$ are $(1+\sqrt{\gamma^{-1}})^2$ and
$(1-\sqrt{\gamma^{-1}})^2$. 

When
$\gamma =1$, and the entries are i.i.d. sub-Gaussian, the maximal
singular value is concentrated around $2$, but the minimal one
is $\min\{n^{-1},d^{-1}\} ( \max \{\sqrt{n}-\sqrt{d-1},\sqrt{d}-\sqrt{n-1}\})^2$ was observed in \cite{rudelson2009smallest}, for normal random variables it was first observed in \cite{davidson2001local}.

Fo a  system of linear equations $Ax=b$, when $n \approx d$, it is better to reduce/increase the data/variables (i.e. 
''better'' to have more variables than data). The
condition number associated with the minimum norm solution
$x=A^\dagger b$ is usually much better -- that is closer to $1$ --
than the condition number of a well-determined system with $n=d$, if
the matrix $A$ is random (see for instance \cite{Chen2005ConditionNO,hastie2019surprises}).

\begin{figure}
	\centering
	\includegraphics[trim = 110 240 125 255, width=1\linewidth, clip]{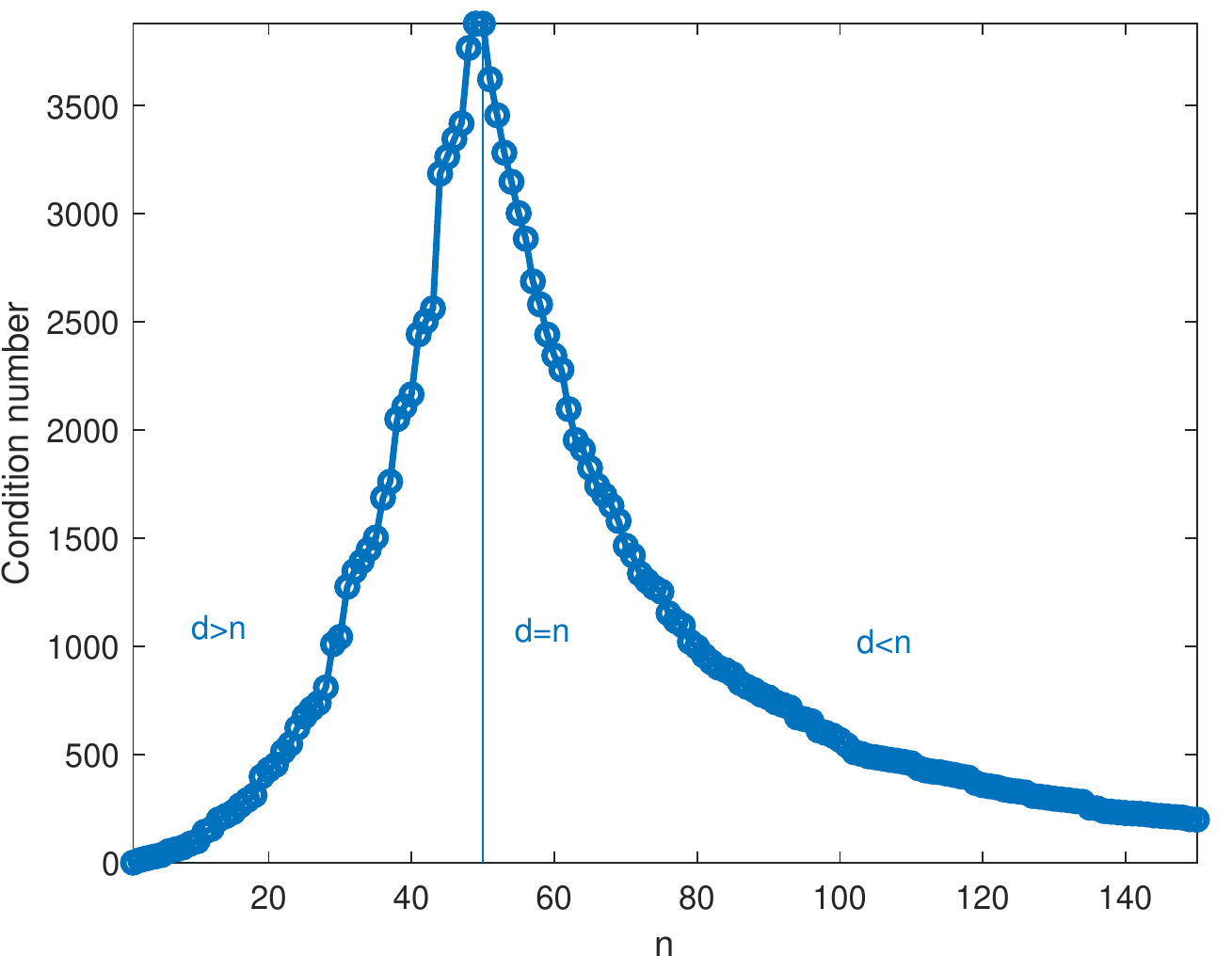}
	\caption{Typical ''double descent" of the condition number of the matrix $K(x_i,x_j)$, where $K(x,x') =  \exp\left(-\frac{||x-x'||^2}{2\sigma^2}\right)$ (radial kernel) and $x_1,\ldots x_n$ are i.i.d. $\mathcal{N}(0,I_{d \times d})$.
		The condition number exhibts the same behavior as in the linear case (here  $\sigma=5$).}
	\label{CondNumberrbf}
\end{figure}

There are  interesting observations for machine learning. The most
obvious is that kernel methods, which are a popular
workhorse in machine learning, {\it do not require regularization in
  order to be well-conditioned}, if the kernel matrices are based on
high dimensional i.i.d data, especially when $\gamma<1$.  This claim
follows from recent results on kernels.  The simplest form of the
kernel matrix $K(x_j,x_i)$ is $K=XX^T$. We consider random matrices whose entries are
$K(x_i^Tx_j)$ with i.i.d. vectors $x_i$ in
$\mathbf{R}^p$ with normalized distribution (in Figure \ref{CondNumberrbf} we consider a
radial kernel $K(||x_i-x_j||^2)$ for which similar arguments are
likely to hold). Assuming that $f$ is sufficiently
smooth and the distribution of $x_i$'s is sufficiently nice, El Karoui
\cite{2010arXiv1001.0492E} showed that the spectral distributions of
kernel dot-product matrices $K(x_i,x_j)=f(\frac{1}{d} XX^T)$ behave as if $f$ is
linear in the Marchenko--Pastur limit. In fact, El Karoui showed that
under mild conditions, the kernel matrix is asymptotically equivalent
to a linear combination of $XX^T$, the all-1’s matrix, and the
identity, and hence the limiting spectrum is Marcenko-Pastur. As a
consequence, the claims about the condition number of a random matrix
$A$ also apply to kernel matrices with random data, see Figure \ref{CondNumberrbf}.

More intriguing is the fact that the behavior of the condition number
of $K^\dagger$ is similar to the {\it double descent} behavior of the
test error by linear and kernel interpolants, which after pioneering
work by Belkin (\cite{Belkin15849}, see also
\cite{2017arXiv171003667A}) has recently attracted much attention
\cite{DBLP:journals/corr/abs-1903-07571,2018arXiv18,Belkin15849,2019arXiv190805355M,
  2018arXiv181211167R,2018arXiv180800387L,2019arXiv190308560H}. We
will address the key role of stability for the theory of machine
learning in a separate paper.

\vskip 0.1in

\subsubsection*{Acknowledgments}
We thank Gill Strang for very useful comments
and for encouraging to publish this note. We are grateful to Felipe Cucker, Misha
Belkin, lorenzo Rosasco, Aleksander Madry and especially Silvia Valle. This material is based upon work
supported by the Center for Minds, Brains and Machines (CBMM), funded
by NSF STC award CCF-1231216, and part by C-BRIC, one of six centers
in JUMP, a Semiconductor Research Corporation (SRC) program sponsored
by DARPA. This research was also sponsored by grants from the National
Science Foundation (NSF-0640097, NSF-0827427), and AFSOR-THRL
(FA8650-05-C-7262).

\bibliographystyle{unsrt}

\bibliography{Boolean} \normalsize

\end{document}